\documentclass[10pt,twocolumn,letterpaper]{article}

\usepackage{iccv}              
\usepackage{arydshln}
\usepackage[accsupp]{axessibility}

\DeclareMathOperator*{\argmin}{arg\,min}

\definecolor{iccvblue}{rgb}{0.21,0.49,0.74}
\usepackage[pagebackref,breaklinks,colorlinks,allcolors=iccvblue]{hyperref}

\def\paperID{1311} 
\def\confName{ICCV}
\def\confYear{2025}

\title{Auto-Regressively Generating Multi-View Consistent Images}

\vspace{-1cm}

\author{JiaKui Hu$^{1,2,3,4}$\thanks{This work was done when they interned in Baidu. Equal contribution.}, Yuxiao Yang$^{5,2}$\footnotemark[1], Jialun Liu$^{2}$\thanks{Corresponding author.}, Jinbo Wu$^{2}$\thanks{Project leader.}, Chen Zhao$^{2}$, Yanye Lu$^{1,3,4}$\footnotemark[2]\\
$^1$Institute of Medical Technology, Peking University Health Science Center, Peking University \\
$^2$Baidu VIS\\
$^3$Biomedical Engineering Department, College of Future Technology, Peking University \\
$^4$National Biomedical Imaging Center, Peking University\\ 
$^5$Tsinghua University\\
{\tt\small jkhu29@stu.pku.edu.cn, yangyuxi23@mails.tsinghua.edu.cn, liujialun95@gmail.com,}\\ 
{\tt\small yanye.lu@pku.edu.cn} 
}

\begin{document}
\maketitle
\begin{abstract}
Generating multi-view images from human instructions is crucial for 3D content creation. The primary challenges involve maintaining consistency across multiple views and effectively synthesizing shapes and textures under diverse conditions. In this paper, we propose the Multi-View Auto-Regressive (\textbf{MV-AR}) method, which leverages an auto-regressive model to progressively generate consistent multi-view images from arbitrary prompts. Firstly, the next-token-prediction capability of the AR model significantly enhances its effectiveness in facilitating progressive multi-view synthesis. When generating widely-separated views, MV-AR can utilize all its preceding views to extract effective reference information. Subsequently, we propose a unified model that accommodates various prompts via architecture designing and training strategies. To address multiple conditions, we introduce condition injection modules for text, camera pose, image, and shape. To manage multi-modal conditions simultaneously, a progressive training strategy is employed. This strategy initially adopts the text-to-multi-view (t2mv) model as a baseline to enhance the development of a comprehensive X-to-multi-view (X2mv) model through the randomly dropping and combining conditions. Finally, to alleviate the overfitting problem caused by limited high-quality data, we propose the ``Shuffle View" data augmentation technique, thus significantly expanding the training data by several magnitudes. Experiments demonstrate the performance and versatility of our MV-AR, which consistently generates consistent multi-view images across a range of conditions and performs on par with leading diffusion-based multi-view image generation models. The code and models are released \href{https://github.com/MILab-PKU/MVAR}{here}.
\end{abstract}

    
\section{Introduction}
\label{sec:intro}

Generating multi-view images from human instructions, \textit{e.g.}, texts, reference images, and geometric shapes, represents a crucial endeavor with profound implications in domains such as 3D content creation, robotic perception, and simulation. The challenge lies in the development of models capable of generating multi-view consistent images while addressing diverse prompts in a unified architecture.

\begin{figure}
\centering
\includegraphics[width=\linewidth]{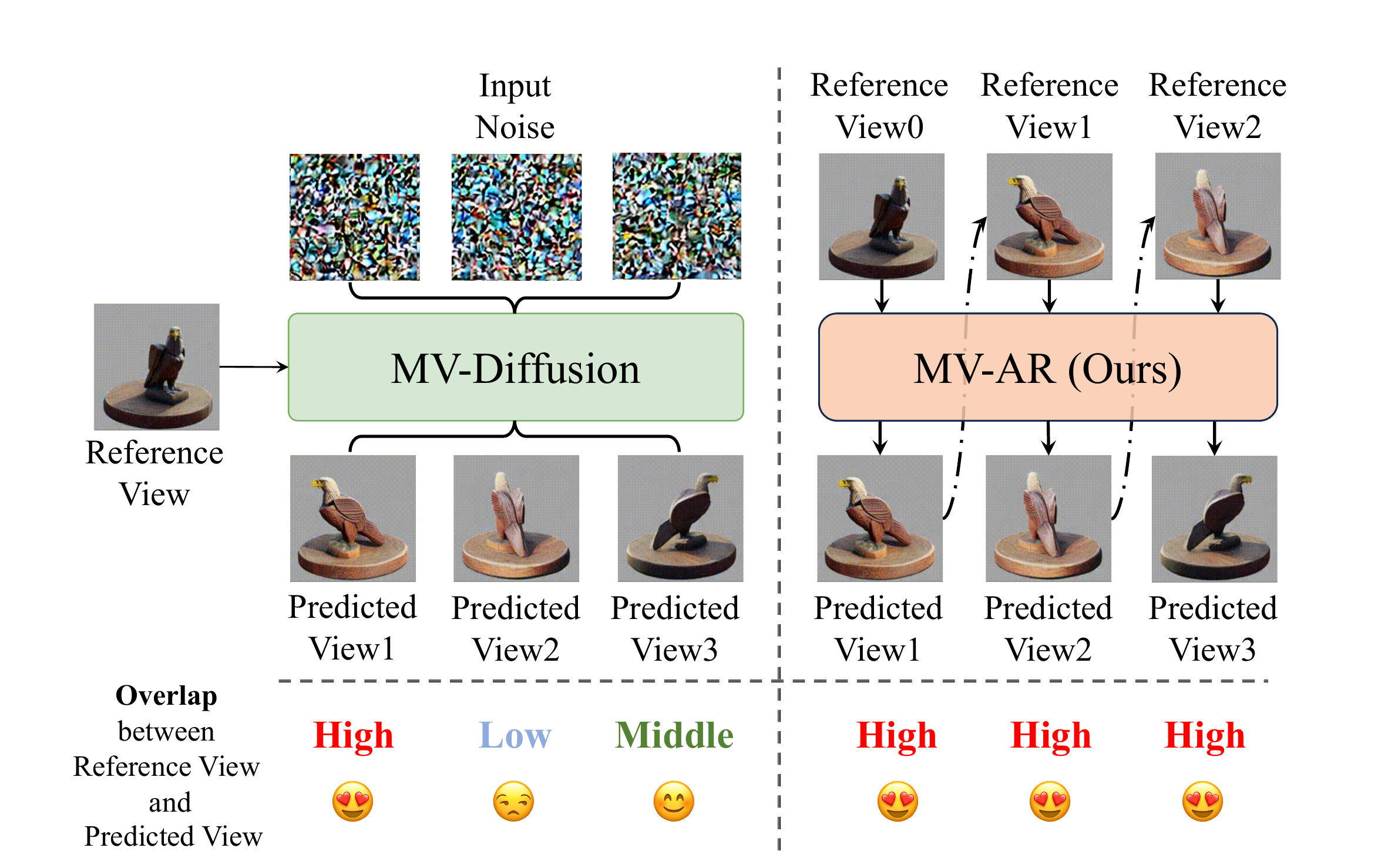}
\vspace{-7mm}
\caption{Diffusion-based multi-view image generation methods use a specific reference view for predicting subsequent views, which becomes problematic when overlap between the reference view and the predicted view is minimal, affecting image quality and multi-view consistency. Our MV-AR addresses this by using the preceding view with significant overlap for conditioning.}
\vspace{-3mm}
\label{fig:overview}
\end{figure}

Recent multi-view image generation approaches~\cite{shi2023mvdream,liu2023zero,shi2023zero123plus,liu2023syncdreamer,long2024wonder3d,li2024era3d} concentrate on leveraging the image generation priors of the pre-trained diffusion model~\cite{rombach2022high}. With the advancement of video diffusion models, some methods have evolved to utilize pre-trained video diffusion models~\cite{blattmann2023stable} for multi-view images with exemplary outcomes~\cite{zuo2024videomv}. As shown in Figure~\ref{fig:overview}, these methods utilize text or images as conditions, facilitating simultaneous synthesis across multiple views. However, when synthesizing images from distant views, the overlap between reference and target images decreases significantly, weakening the effectiveness of reference guidance. In extreme cases, such as generating a back-view image from a front-view reference, the visual reference information becomes nearly negligible due to minimal overlapping textures. Consequently, this insufficient reference between distant views severely compromises the multi-view consistency.

To address this limitation, we propose employing Auto-Regressive (AR) generation for multi-view image generation. As shown in Figure~\ref{fig:overview}, in AR-based generation, the model leverages information derived from the preceding $n-1$ views as conditions to generate the $n$-th view, allowing the model to utilize information derived from previously generated views. This generation process is highly consistent with how humans observe 3D objects. For instance, in scenarios where a back-view is generated from a front-view reference, the AR generation model extracts adequate and pertinent reference from the preceding views. Inspired by this concept, we propose the Multi-View Auto-Regressive (\textbf{MV-AR}) model. MV-AR establishes a novel framework for generating multi-view consistent images, which uses all previous view information to predict patches of subsequent views, thus facilitating the model's ability to capture efficient reference for distant views.

However, new framework introduces new practical issues. This study is dedicated to the examination of the issues~\footnote{This study focuses on addressing the AR model's issues associated with generating multi-view images. The resolution of fundamental issues inherent to AR models, including their unidirectional nature and discrete encoding, is designated for future work.} associated with the application of AR models in the context of multi-view image generation and proposes solutions thereto. We delineate three main issues that inhibit the effective development of AR models within multi-view image generation: \textit{insufficient conditions}, \textit{limited high-quality data}, and \textit{cumulative error}. We systematically address these problems from two distinct perspectives: \textit{architectures} and \textit{training}. We devise effective methods for conditional injection that apply to text~\cite{shi2023mvdream,richardson2023texture}, camera pose~\cite{kant2024spad}, reference image~\cite{liu2023zero,liu2023syncdreamer}, and shape~\cite{richardson2023texture,xiong2025texgaussian} in our MV-AR. Specifically, we practically demonstrate that the in-context image condition is not suitable for the current AR model. We design a precise Image Warp Controller (IWC) module that extracts features of the overlap between the current view and the previous views. IWC inserts condition token-by-token, thereby ensuring accurate image control. Next, we progressively drop or combine multi-modal conditions, so that the model can handle multi-modal conditions simultaneously after training. Finally, to address the challenge of limited training data, we formulate a data augmentation termed ``Shuffle Views" and a progressive learning strategy to enrich the training data and mitigate the risk of overfitting. Utilizing the aforementioned solutions, we formulate a series of guidelines for the AR Model in multi-view image generation and establish a new baseline for its application.


In summary, our contributions to the multi-view image generation field include:

\begin{itemize}
\item \textbf{Auto-Regressively generating multi-view images}. We first introduce the Auto-Regressive (AR) framework in the multi-view image generation task. AR can effectively capture the information derived from all the preceding views as conditions, thereby improving multi-view consistency across far-away views. 
\item \textbf{Architectures and training design}. We investigate the unique challenges faced by AR models in multi-view image generation and implement targeted designing with respect to both architecture and training strategies. Based on this, we establish a robust baseline for AR-based multi-view generation approaches.
\item \textbf{Performance and versatility.} We significantly narrow the performance gap between AR-based and diffusion-based multi-view generation methods. We also develop the first unified (X2mv) multi-view generation model, capable of handling various conditions such as text, image, and shape synchronously.
\end{itemize}

\section{Related work}
\subsection{Multi-view Diffusion Models}
To generate consistent multi-view images, MVDream~\cite{shi2023mvdream} proposes to generate multi-view images conditioned on a given text prompt via a cross-view attention mechanism. Zero123++~\cite{shi2023zero123plus} tiles multi-view images into a single frame, performs a single pass for multi-view generation, and extends the pre-trained self-attention layer to facilitate cross-view information exchange, a technique also employed in Direct2.5~\cite{lu2024direct2} and Instant3D~\cite{li2023instant}. Syncdreamer~\cite{liu2023syncdreamer} integrates multi-view features into 3D volumes, conducting 3D-aware fusion in 3D noise space. Wonder3D~\cite{long2024wonder3d} introduces a multi-view cross-domain diffusion model that enhances cross-domain and cross-view information exchange through distinct attention layers. However, all of the aforementioned methods share the same core idea: modeling 3D generation using a multi-view joint probability distribution via an additional attention mechanism, which is inherently unnatural and fails to capture the progressive nature of novel view generation. Moreover, diffusion-based multi-view generation methods~\cite{shi2023mvdream,shi2023zero123plus,liu2023syncdreamer,long2024wonder3d,liu2025texgarment,liu2024texoct} often require significant architectural modifications when the conditioning type changes. In contrast, we propose a method that generates multi-view images in an auto-regressive manner and incorporates multi-modal conditions in a unified framework.

\subsection{Autoregressive Visual Generation} Pioneered by PixelCNN~\cite{van2016conditional}, researchers have proposed generating images as sequences of pixels. Early research, including VQVAE~\cite{van2017neural} and VQGAN~\cite{esser2021taming}, quantizes image patches into discrete tokens and employs transformers to learn Auto-Regressive (AR) priors, similar to language modeling~\cite{brown2020language}. To further improve reconstruction quality, RQVAE~\cite{lee2022autoregressive} introduces multi-scale quantization, while VAR~\cite{tian2025visual} reformulates the process into next-scale prediction, significantly enhancing sampling speed. Parallel efforts have been dedicated to scaling up AR models for text-conditioned visual generation tasks~\cite{han2024infinity}. As for the 3D area, although some preliminary works~\cite{nash2020polygen,siddiqui2024meshgpt} employ AR models to directly generate vertices, faces of meshes, they are limited to the geometry domain and struggle to scale to general 3D object datasets~\cite{deitke2023objaverse}. Unlike the aforementioned methods, our model is based on a pre-trained text-to-image AR model~\cite{sun2024autoregressive}, leveraging its strong generation prior and extending it to multi-view generation tasks conditioned on diverse input types.
\section{Methods}


\subsection{Auto Regressive Model in Multi-View}

\textbf{Formulation.} The vanilla auto-regressive (AR) model is designed to infer the distribution of a long sequence. Specifically, given a sequence ${x}$ of length $T$, the AR model seeks to derive the distribution of the ${x}$ according to the following formula:

\begin{equation}\label{eq:vanilla_ar}
p(x_1, x_2, \dots, x_T) = \prod_{t=1}^{T} p(x_t | x_{<t}),
\end{equation}

\noindent where $x_i$ denotes the $i$-th datapoint in sequence $x$ and $x_{<t} = (x_1, x_2, \dots, x_{t-1})$ denotes the vector of random variables with index less than $t$. Training an AR model $p_{\theta}$ involves optimizing $p_{\theta} (x_t | x_{<t})$ over large-scale image sequences. 

In image generation tasks, these sequences are generated by vision tokenizers~\cite{van2017neural,esser2021taming,lee2022autoregressive}. The encoder $\varepsilon(\cdot)$ extracts the high-dimensional features $f \in \mathbb{R}^{h \times w \times D}$ of an image $I$. Subsequently, $f$ is flattened into discrete 1d tokens $q \in [V]^{h \times w}$. The quantizer $\mathcal{Q}$ generally comprises a downloadable codebook $Z \in \mathbb{R}^{V \times C}$ that contains $C$ vectors. In the quantization process $q = \mathcal{Q}(f)$, each feature vector $f (i,j)$ is assigned to the code index $q^{(i-1)*w+j}$ of its closest entry in the codebook:



\begin{equation}
q^{(i-1)*w+j} = \left( \argmin_{v \in [V]} \| \text{lookup}(Z, v) - f^{(i,j)} \|_2 \right) \in [V],
\end{equation}

\noindent where the term lookup$(Z, v)$ refers to the retrieval of the $v$-th vector in the codebook $Z$. The notation indexed as superscript $(i-1)*w+j$ of $q$ signifies that $q$ is flattened, whereby the data originally located at the 2d coordinate $(i, j)$ are transformed into its 1d counterpart $(i-1)*w+j$.

\noindent \textbf{Reformulation.} Under the multi-view image generation task, the construction of the sequences $q$ is different from that of a single 2D image. Specifically, we use 2D VQVAE~\cite{sun2024autoregressive} to extract sequences of $N$-view images' features $f=(f_1, f_2, \cdots, f_N)$:

\begin{equation}\label{eq:chain}
\begin{aligned}
q&^{(n-1)*h*w+(i-1)*w+j} = \\
&\left( \argmin_{v \in [V]} \| \text{lookup}(Z, v) - f_n^{(i,j)} \|_2 \right) \in [V],
\end{aligned}
\end{equation}

\noindent where $n$ means the $n$-th view.


\noindent \textbf{Issues of AR model in multi-view image generation.} The aforementioned training sequences facilitate the model's capacity to ensure effective reference across distant views, thus generating consistent multi-view images. However, this framework introduces several issues:

\begin{itemize}
\item \textbf{\textcolor{red}{Issue 1}: Insufficient conditions}. The task of generating multi-view images necessitates that the model adeptly extract features from various conditions and generate multi-view images that maintain consistency with the given conditions. The methods for conditional injection in AR models have not been extensively studied, thereby complicating the efficient utilization of external conditions, \textit{e.g.}, camera poses, reference image, and geometric shape. 
\item \textbf{\textcolor{blue}{Issue 2}: Limited high-quality data}. AR models have been empirically shown to require a substantial volume of high-quality data (such as billions of texts~\cite{achiam2023gpt}) to achieve saturated model training without overfitting. However, the challenges associated with the collection of 3D objects, coupled with the paucity of high-quality multi-view images, significantly hinder MV-AR training adequacy.
\item \textbf{Issue 3: Cumulative error}. Within AR generation, the sequence of images from subsequent $n-1$ views serves as conditions to generate the image in the $n$-th view. In some cases, there exists a low quality image, labeled view $m$ ($m < n$). It is expected to serve as a reliable conditional reference for the subsequent generation of views. However, because of the low quality of the $m$-th view, it fails to provide effective guidance for the target views, leading to a cumulative error in AR generation.
\end{itemize}

These practical limitations call for a rethinking of AR models in the context of multi-view image generation.


\begin{figure*}
\centering
\includegraphics[width=\linewidth]{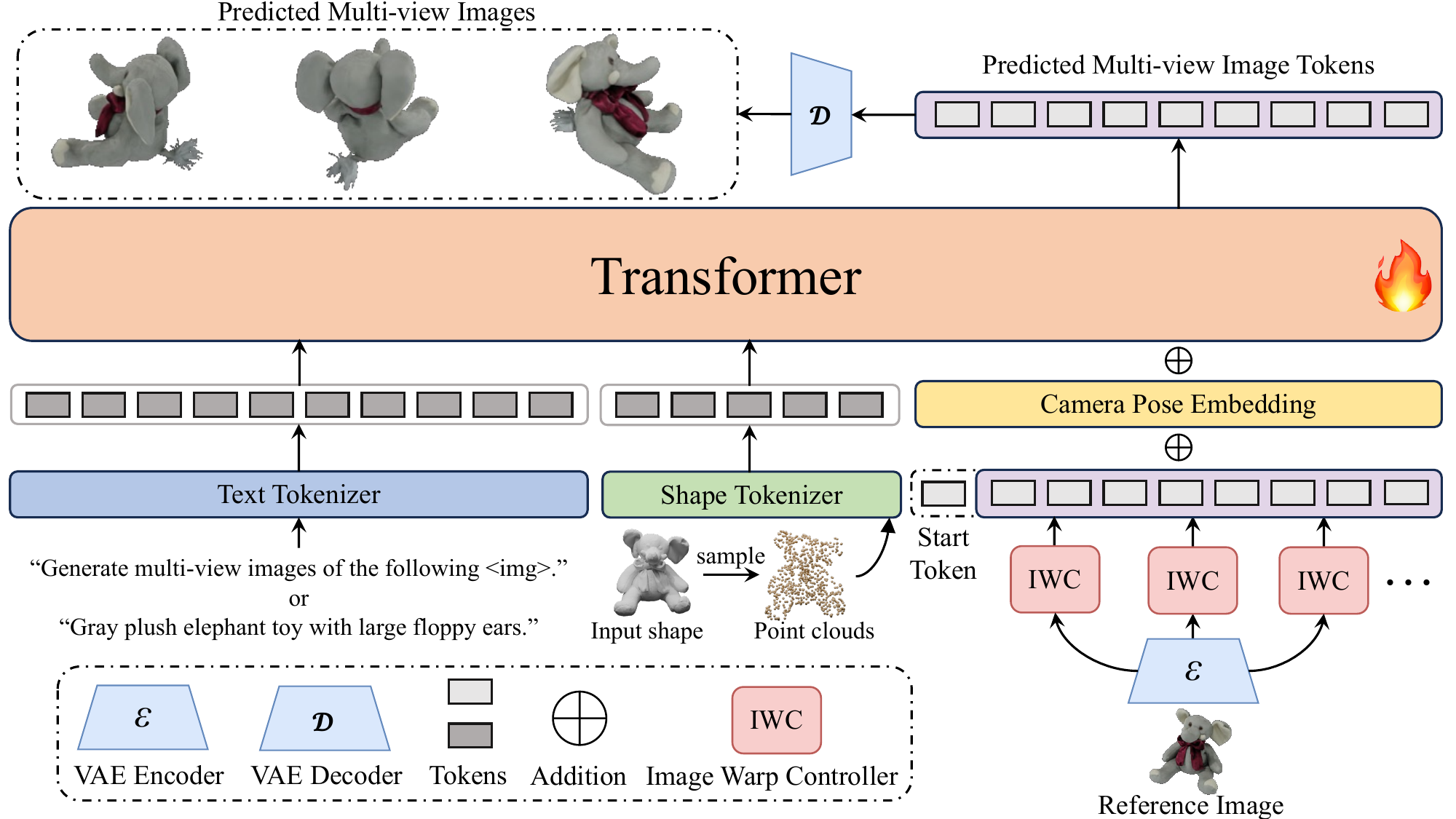}
\vspace{-8mm}
\caption{The overall pipeline of out MV-AR. The text and shape conditions are concatenated before the start token as the context. The text condition can either expect the model to generate multi-view images following other conditions, such as image, or describe the target object. The start token signals the model to begin generating multi-view images. Then, camera pose and image conditions are integrated. The camera pose serves as the shift position embedding, using its angular data to guide the generation of the specific view. After warping by IWC, the image conditions are added token by token within the model. It should be noted that our MV-AR can accommodate these multi-modal conditions simultaneously after progressive learning.}\label{fig:pipeline}
\vspace{-3mm}
\end{figure*}

\subsection{Architectures}

In this subsection, we focus on solving \textbf{\textcolor{red}{Issue 1}} from the perspective of architectures.

\noindent \textbf{Transformer.} Consistent with most AR models, we use Transformer~\cite{vaswani2017attention} to learn auto-regression. Our Transformer is largely based on Llama~\cite{touvron2023llama}, applying pre-normalization using RMSNorm~\cite{zhang2019root}, SwiGLU~\cite{shazeer2020glu} activation function, and AdaLN~\cite{peebles2023scalable}.

\noindent \textbf{Text Condition.} To integrate the text condition into AR models, we use FLAN-T5 XL~\cite{chung2024scaling} as the text encoder, the encoded text feature is projected by an additional MLP and is used as prefilling token embedding in AR models. 

It is crucial to note that, with this in-context text conditions, the Self-Attention (SA) mechanism in the Transformer concurrently integrates both image and text modalities. In situations where there is a misalignment between these modalities, the text tokens might be disturbed by subsequent image tokens, thereby impeding the model's capacity to be effectively directed by the text. Consequently, we develop Split Self-Attention (SSA), a mechanism that harmonizes the efficacy of SA during in-context scenarios with that of cross-attention. 

\begin{equation}\label{eq:ssa}
\begin{aligned}
X_{in} &= \text{Concat}(X_{text}, X_{image}); \\
O_{text}, O_{image} &= \text{Chunk}(\text{SA}(X_{in}), [N_{text}, N_{image}]); \\
\text{SSA}(X_{in}) &= X_{in} + \text{Concat}(0\cdot O_{text}, O_{image}), \\
\end{aligned}
\end{equation}

\noindent where $X_{text} \in \mathbb{R}^{N_{text} \times D}$ is the text feature and $X_{image} \in \mathbb{R}^{N_{image} \times D}$ is the image feature. $\text{Concat}(\cdot)$ concatenates features in token dimension, while $\text{Chunk}(\cdot)$ is the inverse operation of $\text{Concat}(\cdot)$.

As shown in Eq.~\ref{eq:ssa}, SSA is achieved by forcing the output of SA at the specified condition token to zero. This approach ensures that text tokens remain unaffected by ensuing image tokens, while simultaneously maintaining the model's ability to process information pertaining to the image modality. Experiments show that SSA can improve the correspondence between the predicted multi-view images and text condition, such as the CLIP Score~\cite{hessel2021clipscore}.

\noindent \textbf{Camera Condition.} To condition the camera pose, we use the Plücker-Ray Embedding $r \in \mathbb{R}^{(h*w) \times 6}$ that shares the same height and width as the image features $f$ following~\cite{kant2024spad}. It encodes the origin $o$ and direction $d$ of the ray at each spatial location.

\begin{equation}
r_{i,j} = (o \times d, d),
\end{equation}

\noindent where $\times$ is the cross product.

Similarly to image features $f$, in our MV-AR, Plücker ray $r$ is also required to produce a sequence as prescribed by Eq.~\ref{eq:chain}. Given that $r$ represents comprehensive angular information and effectively conveys the positional information of tokens across various views within the entire sequence, we incorporate this ray as the Shift Position Encoding (SPE). Specifically, SPE modifies Eq.~\ref{eq:vanilla_ar} to:

\begin{equation}~\label{eq:spe}
p(x_1, x_2, \dots, x_T) = \prod_{t=1}^{T} p(x_t | (x_0, x_{<t})) + r_{\leq t}),
\end{equation}

\noindent where $x_0$ serves as the start token before the pre-generation image tokens as shown in Figure~\ref{fig:pipeline}.

In Eq.~\ref{eq:spe}, $x_0$ and $x_{<t}$ construct a sequence of length $t$, aligning with the sequence length of the camera condition $r_{\leq t}$, thus allowing for the direct execution of information fusion via the addition operation.

The reason for this shift lies in that $r$ can tell the model which view or patch the next token belongs to~\cite{kant2024spad}. Using $r$, the model can utilize this potential physical angle information to provide precise physical position data for each token within the chains of view, thus enhancing the accuracy of the model predictions for subsequent tokens.

\noindent \textbf{Image Condition.} Alongside the in-context conditioning aligned with the text condition, we develop the Image Warp Controller (IWC) to incorporate fine-grained image conditions into AR models. IWC uses the present camera poses $r$ along with the features $X_{ref}=\varepsilon(I_{ref})$ from the reference view to forecast the features of the overlapped contents and textures $X_{IWC}$ between the current and reference views, subsequently integrating them into the network in a residual fashion. IWC can be formulated as follows.

\begin{equation}
X_{IWC} = \text{FFN}(\text{CA}(\text{SA}(X_{ref}), r)),
\end{equation}

\noindent where CA means the Cross Attention while FFN denotes the Feed-Forward Network.

In IWC, we do not use high-level image conditions, \textit{e.g.}, CLIP~\cite{radford2021learning} or DINO~\cite{caron2021emerging,oquab2023dinov2}, as an image condition. Employing low-level features that more closely approximate the intrinsic attributes of color and texture can facilitate the generation of images with detailed consistency~\cite{caireversible,hu2025universal}.

\noindent \textbf{Shape Condition.}
The inherent one-to-many ambiguity in 3D generation often precludes precise control using text or image inputs alone (\textit{e.g.}, an image cannot fully constrain the underlying 3D shape). However, AR models enable the injection of global shape priors as additional guidance through pre-filled token embeddings, addressing this limitation. Specifically, we adopt point clouds augmented with positional embeddings and normal maps as shape conditions, sampled from 8,192 surface points on the input mesh $M$. We then leverage a pre-trained shape encoder~\cite{zhao2023michelangelo} to map the 3D point cloud into a fixed-length latent token sequence. To enable geometric shape control within the model, the sequence of shape tokens is strategically placed between the sequence of text tokens and the start token, as shown in Figure~\ref{fig:pipeline}.

\subsection{Training}

In this subsection, we focus on solving \textbf{\textcolor{blue}{Issue 2}} and \textbf{Issue 3} from the perspective of training. We examine approaches to address it from three critical perspectives: data augmentation and training strategy.

\noindent \textbf{Loss Function.} The AR model generates the conditional probability $p(q_t | q_{<t} )$ of word $q_t$ at each position $t$. The loss is the average of the negative log-likelihoods over all vocabulary positions:

\begin{equation}\label{eq:loss}
\mathcal{L}_{ar} = -\frac{1}{T} \sum_{t=1}^{T} \log p(q_t | q_{< t}).
\end{equation}

Through optimization of Eq.~\ref{eq:loss}, the model studies the transformation process from the previous $t-1$ tokens to the $t$-th token within the current sequence. Due to the presence of position encoding, the model may be compelled to memorize the token transformation process on the basis of positional information, thereby potentially compromising its capacity for generalization.

\begin{figure*}
\centering
\includegraphics[width=\linewidth]{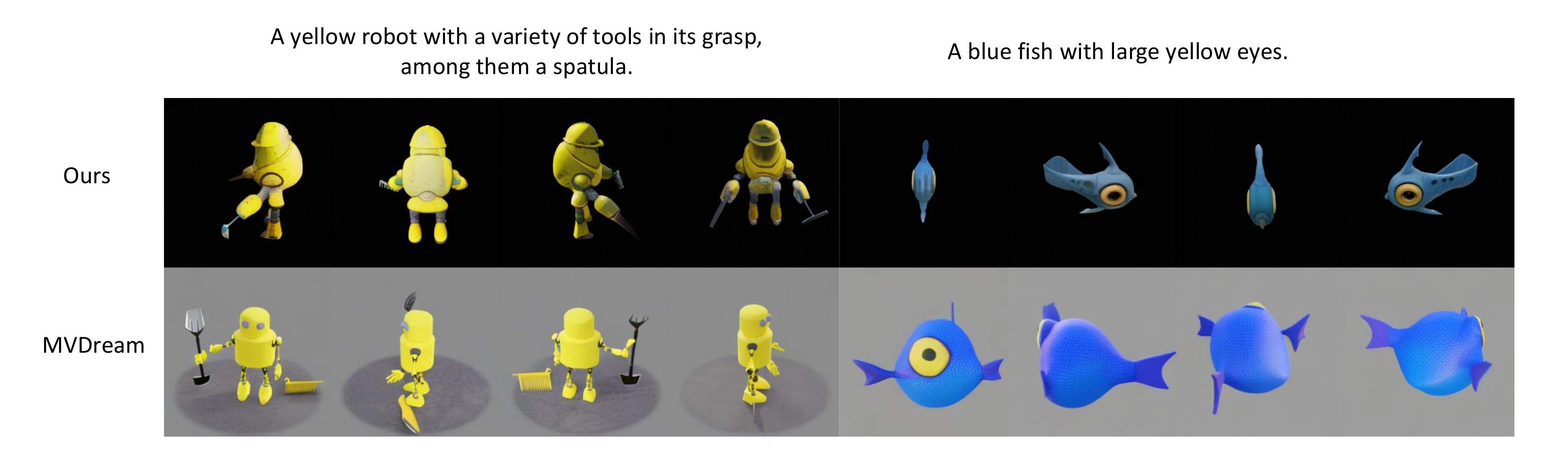}
\vspace{-8mm}
\caption{\textbf{Qualitative study of text to multi-view generation} with diffusion-based method~\cite{shi2023mvdream}.}\label{fig:t2mv}
\vspace{-4mm}
\end{figure*}

\noindent \textbf{Data Augmentation.} In the aforementioned limitations where training data and generalization are limited, data augmentation emerges as an efficient approach to enhance the adequacy of model training. By harnessing the flexibility of perspectives inherent in the multi-view image generation task, we develop the Shuffle Views (ShufV) data augmentation strategy. Upon implementing ShufV, Eq.~\ref{eq:chain} can be reformulated as follows.

\begin{equation}\label{eq:chain_shuffle}
\begin{aligned}
q&^{(S_n-1)*h*w+(i-1)*w+j} = \\
&\left( \argmin_{v \in [V]} \| \text{lookup}(Z, v) - f_n^{(i,j)} \|_2 \right) \in [V],
\end{aligned}
\end{equation}

\noindent where $S$ denotes a random sequence. $S_n$ indicates that the image of the $n$-th view is now utilized as the $S_n$-th view in the training sequence $q$. 

ShufV has the potential to significantly increase the quantity of our training data manifold. In the context of training MV-AR with $n$ views, ShufV can generate $\frac{n(n-1)}{2}$ permutations and combinations of these perspectives. Consequently, a single object or scene can be expanded into $\frac{n(n-1)}{2}$ training sequences, considerably enhancing the diversity and richness of the training data.

It is pertinent to observe that both self-attention and FFN exhibit the permutation equivariance. Consequently, alterations in the sequence order of inputs will induce corresponding changes in the sequence order of the model's intermediate features. To ensure that auxiliary conditions, such as camera poses and reference images, effectively guide the model in generating images in a predetermined sequence, it is imperative to rearrange these conditions. This reordering will ensure alignment of the condition sequence with that of the input sequence.

\noindent \textbf{Discussion on ShufV.} We argue that the ShufV also helps alleviate the part of \textbf{\textcolor{red}{Issue 1}}: the AR model has difficulty exploiting the overlapping conditions between successive and current views. Using ShufV for data augmentation, the order of views is not fixed. Considering that there are two views $\mathcal{A}$ and $\mathcal{B}$ in the input sequence $q$. ShufV enables the AR model to acquire the transformation from view $\mathcal{A}$ to view $\mathcal{B}$ during the training phase, as well as from view $\mathcal{B}$ to view $\mathcal{A}$ after using ShufV. It enhances the IWC's ability to capture the overlap between view $\mathcal{A}$ and view $\mathcal{B}$ with view $\mathcal{B}$ as a reference, and vice versa. Consequently, this permits the model to exploit the overlap conditions between the current view and other views and use them effectively.

\noindent \textbf{Progressive Learning.} Utilizing the aforementioned loss function Eq.~\ref{eq:loss} and the ``ShufV" data augmentation Eq.~\ref{eq:chain_shuffle}, we can successfully train a text-to-multi-view (t2mv) model. This t2mv model employs SSA to analyze text input and efficiently produce multi-view images that correspond to the given text. We use the t2mv model as a baseline to train the X-to-multi-view (X2mv) model to facilitate the versatility of the MV-AR in alternative conditions, such as images and shapes.

During training of the X2mv model, the text condition is randomly dropped, while other conditions are randomly combined, as shown in Figure~\ref{fig:pipeline}. When the text prompt is dropped, it is replaced by a statement that does not pertain to the target image. For example, a command such as ``Generate multi-view images of the following $<$img$>$" may be utilized. In this context, ``$<$img$>$" signifies that a reference image will be combined after the text. In cases where the subsequent element is a geometric shape, ``$<$img$>$" is substituted with ``$<$shape$>$". The probability of a condition dropping and combining increases linearly from 0 to 0.5 as the number of training iterations increases, and remains 0.5 in subsequent training. This escalation is confined to the initial 10k training iterations. This progressive learning allows the model to be influenced by new conditions introduced during training while maintaining a certain level of adherence to the text prompts.

\begin{figure*}
\centering
\includegraphics[width=\linewidth]{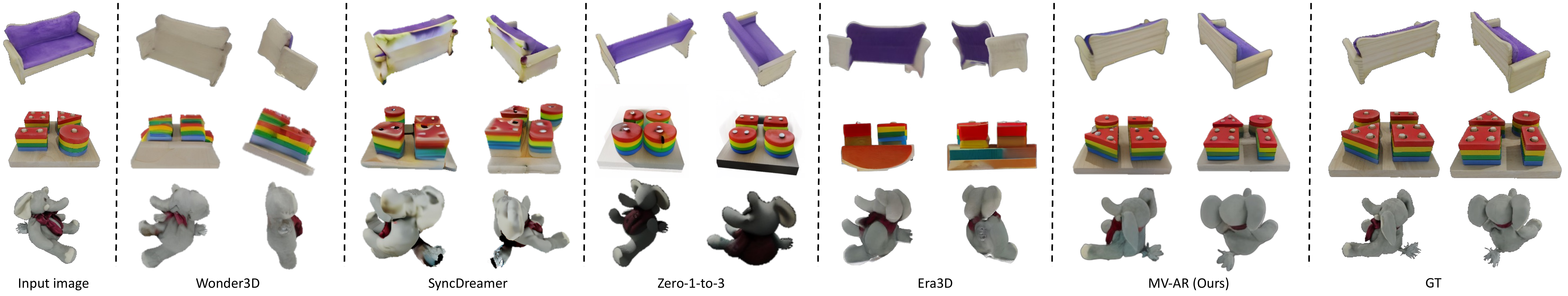}
\vspace{-4mm}
\caption{\textbf{Qualitative comparison on image-conditioned multi-view generation} with diffusion-based method~\cite{long2024wonder3d,liu2023syncdreamer,liu2023zero,li2024era3d}. Era3D~\cite{li2024era3d} generates results in canonical space, so there are certain view differences compared with other methods.}\label{fig:i2mv}
\vspace{-4mm}
\end{figure*}


\section{Experience}

To evaluate the performance and versatility of our MV-AR, we perform a comparative analysis in three specific tasks: (1) text-to-multi-view image generation (refer to Sec.~\ref{sec:t2mv}), (2) image-to-multi-view image generation (refer to Sec.~\ref{sec:i2mv}), and (3) shape-to-multi-view image generation (refer to Sec.~\ref{sec:shape2mv}). In each subsection, we establish ablation studies to evaluate the effectiveness of our designed architectures or training strategies.

\subsection{Text to multi-view}\label{sec:t2mv}

\noindent \textbf{Training details.} Following MVDream~\cite{shi2023mvdream}, we train our MV-AR model on a subset of the objaverse~\cite{deitke2023objaverse} dataset, which comprises 100k high-quality objects. The text instructions are provided by Cap3D~\cite{luo2023scalable}. The training process involves 30k iterations on 16 A800 GPUs with batch-size 1,024 and learning rate $4 \times 10^{-4}$. The optimizer is AdamW with $\beta_1=0.9,\beta_2=0.95$.

\noindent \textbf{Metrics.} The Frechet Inception Distance (FID)~\cite{heusel2017gans} and the Inception Score (IS)~\cite{barratt2018note} are used to assess image quality, while the CLIP-Score~\cite{hessel2021clipscore} is used to assess text-image consistency, with scores averaged across all views.

\noindent \textbf{Results.} 
We adopt the Google Scanned Objects~\cite{downs2022google} as our primary evaluation benchmark. To generate textual descriptions, we render each object from multiple viewpoints and leverage GPT-4V to annotate its visual attributes. A random selection of 30 objects is selected that included the daily items for the animals. For each selected object, an image of size $256 \times 256$ is rendered to serve as an input view. 

\begin{table}[!ht]
\centering
\scalebox{0.9}{
\begin{tabular}{l|c|c|c}
\toprule[0.15em]
Methods & FID $\downarrow$ & IS $\uparrow$ & CLIP-Score $\uparrow$ \\ 
\midrule[0.15em]
MVDream~\cite{shi2023mvdream}$\dag$ & \textbf{141.05} & 7.49 & 28.71 \\
\specialrule{0em}{1pt}{1pt}
\hdashline
\specialrule{0em}{1pt}{1pt}
LLamaGen~\cite{sun2024autoregressive}$\dag$ & 146.11 & 5.78 & 28.36 \\
\textbf{MV-AR (Ours)} & 144.29 & \textbf{8.00} & \textbf{29.49} \\
\bottomrule[0.15em]
\end{tabular}
}
\vspace{-3mm}
\caption{\textbf{Ablation on text-condition module on GSO}. We report FID, IS, and CLIP-Score on the generated multi-view images of 30 GSO objects. $\dag$ means the model is finetuned only on Objaverse subset for fair comparison.}
\vspace{-3mm}
\label{tab:ablation_t2mv}
\end{table}

Table~\ref{tab:ablation_t2mv} illustrates that our MV-AR exhibits a comparable multi-view image quality and text-image consistency consistency. Compared to the diffusion-based method~\cite{shi2023mvdream}, our MV-AR demonstrates equivalent image generation quality while possessing improved image-text consistency, as evidenced by the higher CLIP Score metric attributed to MV-AR. When evaluated against the baseline model LLamaGen~\cite{sun2024autoregressive}, our approach markedly elevates the IS and CLIP-Score metrics due to refinement in processing the text condition from conventional self-attention to the proposed SSA. This demonstrates that our proposed SSA increases the congruence between image and text, as well as the overall quality of generation. The output images are also presented in Figure~\ref{fig:t2mv}. Compared to MVDream~\cite{shi2023mvdream}, the images generated by our MV-AR exhibit a superior multi-view consistency, which is particularly evident in the consistency between the front-view and back-view images. The tool held in the right hand of the yellow robot as generated by MVDream displays significant deformation in front-view and back-view, whereas the yellow robot produced by MV-AR consistently maintains a similar object across all views.

\subsection{Image to multi-view}\label{sec:i2mv}

\noindent \textbf{Training details.} Same as the text-to-multi-view (t2mv) task, the Objaverse subset is used for the training data. Multi-view images, which are arrayed around a circular configuration of objects, are selected as the training images. During training of the 4-view image generation model, the angular difference between each view is set to 90 degrees. Hyper-parameters in training process, such as iterations, learning rate and optimizer, are the same as the t2mv task.

\noindent \textbf{Metrics.} We employ widely used metrics: Peak Signal-to-Noise Ratio (PSNR), Structural Similarity Index Metric (SSIM), and Learned Perceptual Image Patch Similarity (LPIPS)~\cite{zhang2018perceptual}, for the image to multi-view generation task. These metrics are effective in quantitatively assessing the consistency of color and texture between the output image and the corresponding ground-truth image.

\noindent \textbf{Ablation on image condition.} The experimental result in Table~\ref{tab:ablation_i2mv} indicates that cross-attention mechanisms, although widely used within diffusion-based methods~\cite{shi2023mvdream,liu2023syncdreamer}, present significant challenges when applied to AR frameworks. This difficulty is attributed to the base model employed, which lacks inherent image-to-image capabilities. In contrast, the foundational model underpinning diffusion-based methods, \textit{e.g.}, SD2~\cite{rombach2022high} and SDXL~\cite{podellsdxl}, encompasses specific image-to-image functionalities acquired through extensive pre-training. Unlike cross-attention mechanisms, our IWC enables accurate output regulation by incorporating condition token-by-token into the model, thereby facilitating effective control.

\begin{figure*}
\centering
\includegraphics[width=\linewidth]{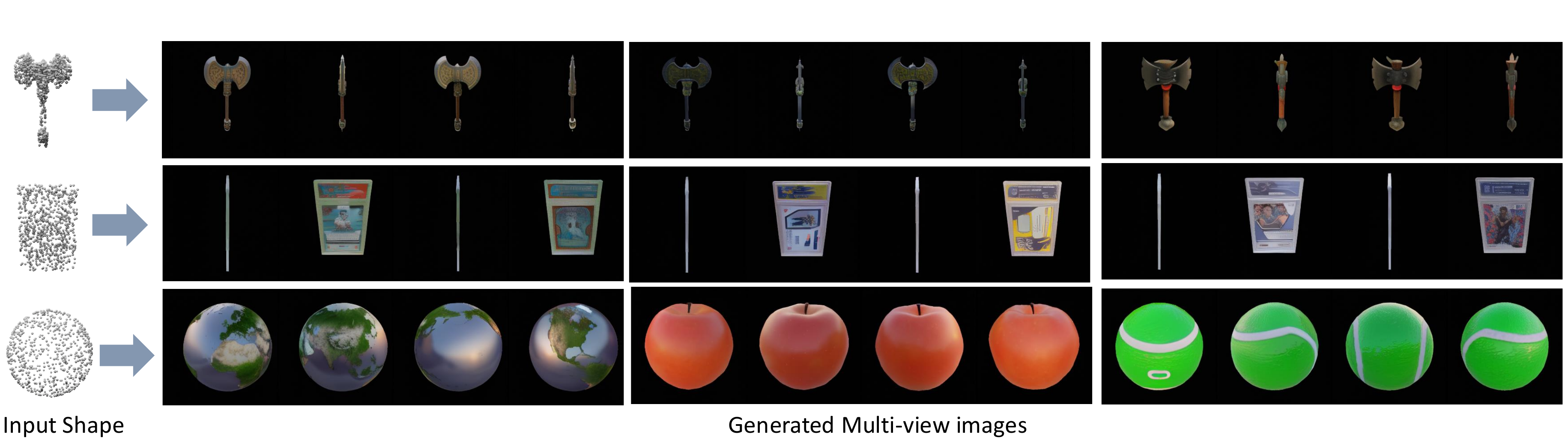}
\vspace{-7mm}
\caption{\textbf{Visualization of our shape-conditioned multi-view generation.} Given a geometric shape, our approach robustly generates multi-view images that are geometrically consistent with it, while ensuring multi-view coherence.}\label{fig:shape}
\vspace{-5mm}
\end{figure*}

\begin{table}[!ht]
\centering
\scalebox{0.9}{
\begin{tabular}{l|c|c|c}
\toprule[0.15em]
Image condition & PSNR $\uparrow$ & SSIM $\uparrow$ & LPIPS $\downarrow$ \\ 
\midrule[0.15em]
In-context & 11.92 & 0.538 & 0.477 \\
Cross Attention & 15.13 & 0.709 & 0.310 \\
\midrule[0.1em]
\textbf{IWC (Ours)} & \textbf{22.99} & \textbf{0.907} & \textbf{0.084} \\
\bottomrule[0.15em]
\end{tabular}
}
\vspace{-3mm}
\caption{\small \textbf{Ablation on image-condition module on GSO.} We report PSNR, SSIM, and LPIPS on the generated multi-view images of 30 GSO objects.}
\label{tab:ablation_i2mv}
\vspace{-5mm}
\end{table}

\noindent \textbf{Evaluation generalization on GSO.} We evaluate the generalization of our MV-AR in image-to-multi-view (i2mv) tasks in Table~\ref{tab:i2mv}. We adopt RealFusion~\cite{melas2023realfusion}, Zero123~\cite{liu2023zero}, SyncDreamer~\cite{liu2023syncdreamer}, Wonder3D~\cite{long2024wonder3d}, and Era3D~\cite{li2024era3d} as comparison methods. Given an input image, these methods generate images of multiple specific views. The quantitative results demonstrate that our MV-AR attains the highest PSNR and the second highest SSIM. These results suggest that our method is more effective in preserving the color and texture fidelity of the reference image, leading to a more consistent generation of multiple views. The LPIPS of our MV-AR is positioned third among all methods, this result indicates that the MV-AR may be overly stringent in enforcing the consistency of low-level features between the generated and reference images, which consequently diminishes the perceptual quality~\cite{zhang2018perceptual}.

\begin{table}[!ht]
\centering
\scalebox{0.9}{
\begin{tabular}{l|c|c|c}
\toprule[0.15em]
Image condition & PSNR $\uparrow$ & SSIM $\uparrow$ & LPIPS $\downarrow$ \\ 
\midrule[0.15em]
Realfusion~\cite{melas2023realfusion} & 15.26 & 0.722 & 0.283 \\
Zero123~\cite{liu2023zero} & 18.93 & 0.779 & 0.166  \\
SyncDreamer~\cite{liu2023syncdreamer} & 19.89 & 0.801 & 0.129  \\
Wonder3D~\cite{long2024wonder3d} & 22.82 & 0.892 & \textbf{0.062}  \\
Era3D~\cite{li2024era3d} & 22.73 & \textbf{0.911} & 0.071 \\
\midrule[0.1em]
\textbf{MV-AR (Ours)} & \textbf{22.99} & 0.907 & 0.084 \\
\bottomrule[0.15em]
\end{tabular}
}
\vspace{-3mm}
\caption{\small \textbf{Quantitative results on image-conditioned multi-view synthesis on GSO}. We report PSNR, SSIM, and LPIPS on the generated multi-view images of 30 GSO objects.}
\label{tab:i2mv}
\vspace{-3mm}
\end{table}

A qualitative comparative analysis of our method against Wonder3D~\cite{long2024wonder3d} and SyncDreamer~\cite{liu2023syncdreamer} is presented in Figure~\ref{fig:i2mv}. We illustrate the superiority of autoregressive multi-view image generation over diffusion-based simultaneous multi-view generation through a scenario where generating back-view images from the front view. MV-AR effectively utilizes the cumulative information from all preceding viewpoints, as opposed to relying on a singular view. This progressive generation allows for a comprehensive reference during generation, leading to better qualitative results.

\subsection{Shape to multi-view}\label{sec:shape2mv}

\noindent \textbf{Training details.} The training data and hyper-parameters used are consistent with those used in the image-to-multi-view task. Furthermore, we implement our proposed progressive learning approach to enhance the model's ability in multi-modal conditional processing.

\noindent \textbf{Results.} To validate the efficacy of our shape-conditioned multi-view generation framework, we fix the shape token and generate multi-view outputs across multiple trials, as illustrated in Figure~\ref{fig:shape}. Our method consistently produces outputs that not only adhere to the input shape constraints with high precision but also exhibit semantically plausible and diverse textures.

\subsection{Ablation study}

\begin{table}[!ht]
\centering
\scalebox{0.9}{
\begin{tabular}{c|c|c}
\toprule[0.15em]
& FID $\downarrow$ / IS $\uparrow$ & PSNR $\uparrow$ / SSIM $\uparrow$ / LPIPS $\downarrow$ \\
\midrule[0.1em]
w/o SPE & 147.29 / 7.26 & 21.30 / 0.843 / 0.118 \\
w/o ShufV & 173.51 / 4.77 & 18.27 / 0.778 / 0.194 \\
\midrule[0.1em]
\textbf{MV-AR (Ours)} & \textbf{144.29} / \textbf{8.00} & \textbf{22.99} / \textbf{0.907} / \textbf{0.084} \\
\bottomrule[0.15em]
\end{tabular}
}
\vspace{-3mm}
\caption{\small \textbf{Ablation study of SPE and ShufV}, with the performance on the t2mv and i2mv tasks presented simultaneously.}
\vspace{-3mm}
\label{tab:ablation_spe_shufv}
\end{table}

\noindent \textbf{Effect of SPE camera pose guidance}. We demonstrate that using the camera pose as the shift position embedding is helpful in generating multi-view consistent images.

\noindent \textbf{Effect of ShufV data augmentation}. We show that the ShufV data augmentation strategy can significantly improve the quality of the output images.





\section{Conclusion}\label{sec:conclusion}

In this study, we introduce an approach for auto-regressively generating multiview images. The motivation is to ensure that, during the generation of the current view, the model can extract efficient guidance information from all preceding views, thus enhancing the multi-view consistency. Based on this idea, we develop the MV-AR model, which progressively produces multi-view images based on multi-modal conditions. Specifically, we initially design several conditional injection modules for four conditions: text, image, camera pose, and geometry. Subsequently, we implement progressive learning to enable the MV-AR to manage these conditions concurrently. Finally, we employ ``Shuffle View" data augmentation to expand the dataset, mitigating the inherent overfitting issues associated with the AR model when applied to limited training data. As a result of these novel designing, our MV-AR model achieves generation performance that is comparable to the state-of-the-art diffusion-based multi-view image generation and establishes itself as the first multi-view image generation model capable of handling multi-modal conditions.


\noindent \textbf{Future work.} We will enhance our MV-AR from: \textit{(1) Better tokenizer}. The reason why we do not use 3D VAE is the exchange of information between views during its encoding, which contradicts the core motivation of our study. Therefore, we will focus on improving performance via tokenizing multi-view images using causal 3D VAE. \textit{(2) Unified generation and understanding}. This study employs the AR model to accomplish the multi-view image generation task. In future work, we aim to harness the comprehensive capabilities of AR to unify the processes of multi-view generation and understanding.

\section*{Acknowledgement}
This work was supported financially the Natural Science Foundation of China (82371112, 623B2001), the Science Foundation of Peking University Cancer Hospital (JC202505), Natural Science Foundation of Beijing Municipality (Z210008) and the Clinical Medicine Plus X - Young Scholars Project of Peking University, the Fundamental Research Funds for the Central Universities (PKU2025PKULCXQ008).

{
    \small
    \bibliographystyle{ieeenat_fullname}
    \bibliography{main}
}

\end{document}